\pdfoutput=1

\documentclass[11pt]{article}

\usepackage[]{ACL2023}

\usepackage{times}
\usepackage{latexsym}

\usepackage[T1]{fontenc}

\usepackage[utf8]{inputenc}

\usepackage{microtype}

\usepackage{inconsolata}
\usepackage{linguex}
\usepackage{booktabs}
\usepackage{graphicx}
\usepackage{amssymb}
\usepackage{amsmath}
\usepackage{makecell}
\usepackage{colortbl}
\usepackage{pifont}

\usepackage{titlesec}
\titlespacing*{\paragraph}{0ex}{0.2ex}{1ex}

\title{Entity Tracking in Language Models}

\author{Najoung Kim\thanks{\ \ Equal contribution. Author order was determined by a fair process of letting Cookie the cat pick a colored ball.} \\
  Department of Linguistics \\
  Boston University \\
  \texttt{najoung@bu.edu} \\\And
  Sebastian Schuster${}^*$ \\
  Dept. of Language Science and Technology \\
  Saarland University \\
  \texttt{seschust@lst.uni-saarland.de} \\}

\begin{document}
\maketitle
\begin{abstract}
Keeping track of how states of entities change as a text or dialog unfolds is a key prerequisite to discourse understanding. Yet, there have been few systematic investigations into the ability of large language models (LLMs) to track discourse entities. In this work, we present a task probing to what extent a language model can infer the final state of an entity given an English description of the initial state and a series of state-changing operations. We use this task to first investigate whether Flan-T5, GPT-3 and GPT-3.5 can track the state of entities, and find that only GPT-3.5 models, which have been pretrained on large amounts of code, exhibit this ability. We then investigate whether smaller models pretrained primarily on text can learn to track entities, through finetuning T5 on several training/evaluation splits. While performance degrades for more complex splits, we find that even when evaluated on a different set of entities from training or longer operation sequences, a finetuned model can perform non-trivial entity tracking. Taken together, these results suggest that language models can learn to track entities but pretraining on text corpora alone does not make this capacity surface.
\end{abstract}

\section{Introduction}
A key prerequisite to long-context understanding and generating coherent text is the ability to accurately represent entities as the discourse unfolds \cite[][\textit{i.a.}]{karttunen1976discourse,groenendijk1991dynamic,heim2002file,nieuwland2006peanuts,kamp2011discourse}. For example, consider the following example in the context of a recipe:

\ex. \label{ex:recipe} Put the eggs, sugar, flour, and baking powder in a bowl and mix to form a light batter. Make sure that the final batter does not contain any lumps of flour or sugar.

In order to understand this instruction, several distinct abilities are necessary:

\vspace{0.15em}\noindent \textbf{New discourse entity recognition:} recognizing when new discourse entities are introduced. E.g., \textit{a bowl} introduces a new discourse entity but \textit{the final batter} or \textit{any lumps of ...} does not.

\vspace{0.15em}\noindent \textbf{Coreference resolution:} associating referring expressions with discourse entities. E.g., \textit{a light batter} and \textit{the final batter} refer to the same entity.

\vspace{0.15em}\noindent \textbf{Discourse entity tracking:} tracking the state changes made to each discourse entity. E.g., \textit{the eggs} are put into \textit{the bowl} and \textit{mixed} with the other ingredients.

\begin{figure}[t]
    \centering
    \includegraphics[width=0.5\textwidth]{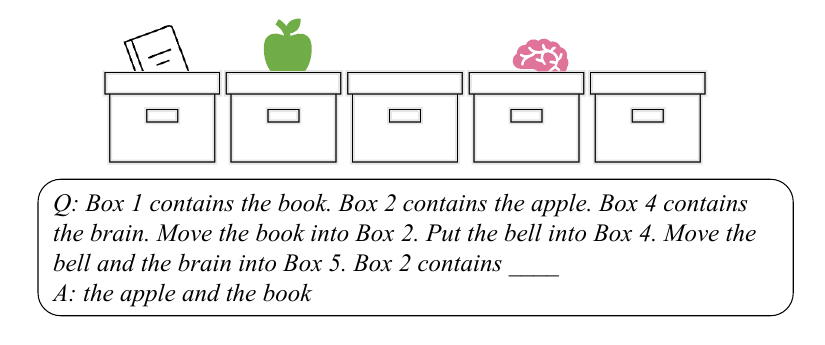}
    \caption{A sketch of our entity tracking task.}
    \label{fig:task}
\end{figure}

There exist many datasets that aim to evaluate these abilities \cite[e.g.,][]{walker2006ace,pradhan-etal-2012-conll,rahman-ng-2012-resolving,weston2015towards,chen-etal-2018-preco,bamman-etal-2020-annotated,uryupina2020annotating} and many NLP models that aim to solve these tasks \cite[e.g.,][]{haghighi-klein-2010-coreference,lee-etal-2011-stanfords,hill2016goldilocks,henaff2017tracking,ji-etal-2017-dynamic,lee-etal-2017-end,bosselut2018simulating,gupta-durrett-2019-effective,gupta-durrett-2019-tracking,aina-etal-2019-entity,toshniwal-etal-2020-learning,wu-etal-2020-corefqa}. In the context of large language models (LLMs), \newcite{tenney-etal-2019-bert}, \newcite{clark-etal-2019-bert}, and \newcite{sorodoc-etal-2020-probing} found that representations of LSTMs and Transformer-based models such as BERT \cite{devlin-etal-2019-bert} do capture coreference relations. \newcite{loaiciga-etal-2022-new} and \newcite{schuster-linzen-2022-sentence} found that pretrained models are able to detect whether noun phrases introduce discourse entities, albeit not fully systematically.

The question of whether LLMs \textit{can track the state of discourse entities}, however, has mostly been indirectly evaluated. \newcite{toshniwal-etal-2022-chess} showed that GPT-2 \cite{radford2019language} can learn to predict valid chess moves based on a compact, nonlinguistic description of previous moves. Similarly, \newcite{li2023emergent} showed that a GPT model trained on Othello can predict valid next moves, and that these predictions are tied to the model's internal representations of the board states. Still, these results do not tell us whether LLMs track state changes expressed in natural language discourses. The most relevant evaluation is \newcite{li-etal-2021-implicit}, where they tested whether model representations encode entity states described in naturalistic text. Using a probing classifier, they found that the states can be decoded from T5 \citep{raffel2020exploring} and BART \citep{lewis-etal-2020-bart} with high accuracy. However, as we show in a reanalysis of their results (Section~\ref{sec:motivation}), they do not provide definitive evidence for entity tracking. Hence, whether LLMs can track entities during the processing of natural language discourse remains an open question.

\paragraph{Contributions} This work attempts to answer this question by developing a task targeted towards evaluating a language model's ability to track state changes of discourse entities (illustrated in Figure~\ref{fig:task}). We use this novel task to evaluate GPT-3 \cite{brown2020language}, \href{https://beta.openai.com/docs/model-index-for-researchers/models-referred-to-as-gpt-3-5}{GPT-3.5},
and Flan-T5 \cite{chung2022scaling} without any finetuning. We find that only models in the GPT 3.5 series, which have been trained on both text and code, are able to perform non-trivial entity tracking. We then show that a smaller language model (T5) can learn to perform non-trivial entity tracking and also demonstrates some capacity to generalize to state descriptions with more operations or with low lexical overlap. Our results suggest that language models can learn to track entities but pretraining on text corpora alone does not make this capacity surface. More broadly, our task can also serve as a useful tool for investigations into emergent world models in LMs \cite[e.g.,][]{li2023emergent,tsai-etal-2023-large}.\footnote{Code and data are available at \url{https://github.com/sebschu/entity-tracking-lms}.} 

\section{Reanalysis of \citet{li-etal-2021-implicit}}
\label{sec:motivation}

We start from examining \citet{li-etal-2021-implicit}, the most relevant work to ours. They adapted two existing datasets, Alchemy \citep{long-etal-2016-simpler} and TextWorld \citep{cote2019textworld}, to test a model's ability to track state changes of an entity. The input to the model is a text description of the initial world state followed by state-changing instructions. Based on this description, the model is expected to identify the correct final state of each entity. For example, for Alchemy, the model receives formulaic descriptions of 7 beakers containing different amounts of colored liquids, followed by instructions that manipulate their contents such as pouring the liquid from one beaker into another, or draining a beaker. Given an input like \ref{ex:alchemy}, the model is expected to recognize that the first beaker has 4 units of brown liquid, the second beaker has 2 units of red liquid, and the third beaker is empty.

\ex. \label{ex:alchemy} \textit{The first beaker has 1 green, the second beaker has 2 red, the third beaker has 3 red. Pour the last red beaker into beaker 1. Mix.}

Using such descriptions, \citet{li-etal-2021-implicit} found that a probing classifier that takes as input the encoding of these descriptions from T5 or BART is able to correctly predict the state of 75--76\% of the entities, suggesting some degree of success on entity tracking.

However, this conclusion becomes questionable when the datasets and the results are scrutinized further. Specifically, we conducted a fine-grained analysis of the success cases of the Alchemy experiment. In this experiment, the state of each beaker was probed after each state-changing instruction. Because each instruction targets at most two beakers (e.g., \textit{pour X into Y}) and there are 7 beakers in total, there is a sparse representation of cases probing a beaker that actually underwent a change in the dataset. Indeed, 62.7\% of all beaker states probed were identical to the initial state, meaning that a simple baseline that always predicts the initial state already achieves 62.7\% accuracy (this is also noted by \citeauthor{li-etal-2021-implicit}). A second potential for shortcuts was the high rate of empty final states (32.4\%).\footnote{This percentage also includes cases where a beaker was already empty in the initial world state description. The percentage of cases where a beaker was empty but contained some liquid in the initial description is 24.9\%.} For these cases, the initial state can often be entirely disregarded, due to the presence of an emptying instruction such as \textit{Drain the fourth beaker}. This instruction alone is sufficient to predict the fourth beaker's final state independent of its initial state. Therefore, such examples are also not best suited to fully assess entity tracking. Given the high prevalence of these two trivial scenarios (87.6\% in total), only 12.4\% of the datapoints can be considered as truly assessing state changes unfolding over a discourse context. If the accuracy is computed on the trivial and non-trivial cases separately, the probing classifier achieves 86.8\% accuracy on trivial cases but only 3.1\% accuracy on non-trivial cases, showing that most of the reported success derives from the trivial cases.

In summary, our reanalysis suggests that the results of \citet{li-etal-2021-implicit} do not provide conclusive evidence for non-trivial state tracking abilities in language models.\footnote{\label{footnote:li}\newcite{li-etal-2021-implicit} also presented two other sets of experiments. See Appendix~\ref{app:implicit-meaning} for details on how the other experiments exhibit similar issues.} However, it remains unclear whether this is due to issues with the setup or a true lack of entity tracking capacity. To this end, we propose a new behavioral evaluation.

\section{Task Design and Dataset}
\label{sec:dataset-task}

\subsection{Desiderata}
\label{subsec:desiderata}

The ability to track entities should be largely \textit{independent of specific linguistic forms}. For a model that can properly track entities, it should not matter whether one talks about beakers or recipes or which specific syntactic constructions are used. This makes it an interesting ability to evaluate in the context of assessing whether and how meaning is represented, since at least classic language models are only trained on forms \cite{bender-koller-2020-climbing}. At the same time, this independence of form and entity states that should hold for true entity tracking poses a challenge for evaluation, since one needs to ensure that the state of entities cannot be predicted from individual lexical items or phrases (such as the word \textit{drain} in the Alchemy dataset, as discussed in Section~\ref{sec:motivation}). Furthermore, language models pretrained on large corpora may have learned common states of entities; for instance, that eggs often end up in a bowl. For these reasons, any task that evaluates entity tracking abilities should conform to the following four desiderata:
\begin{enumerate}
\item The probed states of entities should not follow similar distributional patterns to those that are likely to be present in the pretraining data \cite[see also][]{linzen-2020-accelerate}.
\item Individual words or phrases should not predict by themselves the state of an entity without considering the previous discourse in order. 
\item If any data is used for demonstration, finetuning or training, the training and evaluation data should have little lexical overlap.
\item If any data is used for demonstration, finetuning or training, the task should not be solvable by slot-filling based on observed datapoints.
\end{enumerate}

\noindent These properties cannot be guaranteed with naturalistic datasets such as recipes \cite{kiddon-etal-2015-mise}, science texts \cite{dalvi-etal-2019-everything}, or the Alchemy and TextWorld datasets, which have been previously used to evaluate entity tracking abilities. We therefore programmatically generated datasets for which these properties hold.

\subsection{Dataset}
\label{subsubsec:dataset}

We take inspiration from \citet{winograd1971procedures} and \citet{li-etal-2021-implicit} in designing our data.
Our datasets consist of text descriptions of a particular state of the world followed by a sequence of changes. The worlds contain boxes that can be filled with objects. The objects can be placed inside the box, taken out of the box, or moved from one box to another. We define a world $\mathcal{W}$ as $\mathcal{W} = (O, n, m, e)$ where $O$ is a set of objects, $n$ is the number of boxes, $m$ is the maximum number of objects one box can contain, and $e$ is the expected number of objects in each box in the initial world states. For our datasets, we used $n=7$, $m=3$, $e=2$, and used a set of nouns denoting items that can plausibly fit inside a box (e.g., \textit{book}, \textit{rock}, \textit{brain}; $|O|=100$), selected from a list of words with frequency greater than 27 in the British National Corpus (BNC; \citealt{leech2001word}).

A dataset consists of multiple distinct \textit{scenarios}. A scenario consists of an initial state and a set of operations applied to this initial state. We fixed the number of operations (NumOps) in each scenario to 12. We randomly sampled 2200 scenarios, where the initial state and the 12 operations were both randomly sampled. The sampling process is designed such that only valid operations given the current world state can be sampled. The initial state and the operations were converted into naturalistic descriptions using predefined templates.

\paragraph{Relation to Desiderata} We selected the task of moving objects across boxes because this is a domain where lexical contents of the entities do not offer cues to predict the outcome of state changes (Desideratum~1). We did not include an operation that empties a box that allows the previous discourse to be discarded (Desideratum~2). For Desideratum~2, we furthermore considered experiments using operation descriptions with greater context dependence. Specifically, we tested scenarios with an additional \textit{Move contents of Box N to Box M} operation that does not explicitly mention the object names, and scenarios where object descriptions in the operations can only be fully disambiguated by knowing the current state of a box. For Desideratum~3, we considered experiments where the phrasing of the states and operations differ entirely between demonstration/finetuning and evaluation (see Table~\ref{tab:altforms-comparison}). Finally, for all experiments, we
computed a ``signature'' of every
initial state that indicates the number of objects contained in each box.\footnote{For example, the signature of an initial state in which the first box contains two objects and the rest contains 1 object each would be 2111111.} Then, we ensured that there were no two examples with identical initial descriptions modulo the object names where one appeared in the training split and the other one in the evaluation split. This prevents this task from being solvable via slot-filling (Desideratum~4).\footnote{Additionally, compared to the Alchemy setup, our setup also has a benefit of requiring fewer additional reasoning abilities. The beaker domain in Alchemy requires the model to count and perform simple arithmetic (e.g., adding one unit of liquid followed by adding two units of liquid results in three units of liquid). Moreover, some of the operations in Alchemy require knowledge about how colors are combined (e.g., mixing red and green liquids results in a brown liquid). The boxes domain removes these requirements.} 

\subsection{Task}

We define the entity tracking task as follows. Given a natural language description of the initial state of the world followed by 0--12 state-changing operations, the content of each box at the end of the description must be correctly identified. To evaluate this, we created a test example for each box after each operation. This corresponds to $n \times (\text{NumOps} + 1) $ examples per scenario (91 exs. in our datasets). Each example is formulated in the style of a cloze test. 
That is, the input describes the initial state followed by a sequence of operations, ending in \textit{Box N contains} \_\_. The expected output is the correct set of objects in Box $N$ based on the prefix description. See Appendix~\ref{app:example} for an example.

\section{Experiment 1: In-context Demonstration}
In the first set of experiments, we evaluated pretrained LMs using a small number of in-context demonstrations of the entity tracking task. This provides a way to probe the model without requiring substantial supervision for the task, as well as guiding the model to output the final state in a consistent format that can be automatically assessed.\footnote{For GPT-3.5, we were able to instruct the model to output predictions in a consistent format without any in-context demonstrations, thus making it a true zero-shot experiment. We discuss the results of this experiment in Appendix~\ref{app:additional-results}.}

\subsection{Models}

We used models that are known to support task adaptation via in-context demonstrations:
GPT-3 175B (\texttt{davinci}: \citealt{brown2020language}), GPT-3.5 (\texttt{text-davinci-003}\footnote{\scriptsize\url{https://beta.openai.com/docs/models/gpt-3}}), and Flan-T5 (base and XL: \citealt{chung2022scaling}). The little information that OpenAI revealed about their models\footnote{\scriptsize\url{https://beta.openai.com/docs/model-index-for-researchers}} suggests that \texttt{davinci} is an autoregressive language model primarily trained on text corpora. \texttt{text-davinci-003} was trained on the language modeling objective on a mix of text and code, and additionally tuned with human feedback using reinforcement learning. Flan-T5 is based on T5, a sequence-to-sequence model trained on a denoising objective, that has been further instruction-finetuned on a battery of tasks. This has been shown to promote better responses to instructions, both with and without demonstrations \cite{chung2022scaling}. We evaluated the GPT models through the OpenAI API and the Flan-T5 using the HuggingFace library \citep{wolf-etal-2020-transformers}.  See Table~\ref{tab:in-context-models} for a summary of the models, and Appendix~\ref{app:model-details} for implementation details.

We compared these models against a baseline that randomly outputs 0 to $m=3$ objects from the set of objects that appeared in the same clauses as the box in question. Note that this baseline is much stronger than a fully random baseline that selects outputs from all mentioned objects.

\begin{table}[]
\resizebox{1.0\columnwidth}{!}{
    \centering
    \begin{tabular}{l c c l}
        \textbf{Model} & \textbf{Size} & \textbf{Code?} & \textbf{Additional Training}  \\
        \midrule
         GPT-3 \texttt{davinci} & 175B & \ding{55} & - \\ \arrayrulecolor{lightgray}\hline
          GPT-3 \texttt{davinci-instruct-beta} & 175B & \ding{55} & \makecell[l]{Human demonstrations \\ (finetuning)} \\\hline
         GPT-3 \texttt{text-davinci-001} & 175B & \ding{55} & \makecell[l]{Human demonstrations\\ + highly rated model outputs \\ (finetuning)} \\\hline
         GPT-3.5 \texttt{code-davinci-002} & ? & \ding{51} & - \\\hline
         GPT-3.5 \texttt{text-davinci-002} & ? & \ding{51} & \makecell[l]{Human demonstrations\\ + highly rated model outputs \\ (finetuning)} \\\hline
         GPT-3.5 \texttt{text-davinci-003} & ? & \ding{51} & \makecell[l]{RL on human feedback} \\\hline
         Flan-T5 base & 250M & \ding{55} & \makecell[l]{1.6K tasks + instructions \\ (finetuning)} \\\hline
         Flan-T5 XL & 3B & \ding{55} & \makecell[l]{1.6K tasks + instructions \\ (finetuning)}
    \end{tabular}
    }
    \caption{Summary of the models used for the in-context demonstration experiments.}
    \label{tab:in-context-models}
\end{table}

\subsection{Prompting and Demonstrations}
\label{subsec:prompting}
Our prompts consist of: (a) a general instruction for the task, (b) two examples of the task to demonstrate the expected format, (c) an initial state description followed by a series of operations, and (d) an incomplete sentence \textit{Box N contains \_\_\_} that the model should complete (see Appendix \ref{app:prompts} for full prompts). We used demonstrations that output the state of all boxes at once. However, in early experiments, Flan-T5 frequently only output the state of the first box even when the in-context demonstrations contained final descriptions of all box states. Therefore, for Flan-T5, we adjusted the prompts to output each box individually.\footnote{To verify that this difference in the task format does not underestimate the accuracy of the GPT models, we also conducted an experiment in which we prompted GPT to only output the state of one box at a time. We found that, contrary to the Flan-T5 models, the accuracy of GPT was \textit{lower} when we prompted it to output individual boxes than prompting it to output all boxes, so we can rule out that this difference in the task format is underestimating GPT's performance.}

\begin{table*}[t]
\centering
\resizebox{2\columnwidth}{!}{
   \begin{tabular}{ccc}
       \textbf{Operation} & \textbf{Base} & \textbf{AltForms} \\\toprule
       \texttt{Move} & \textit{Move the car from Box 1 to Box 3.} & \textit{Pick up the furby in Container A and place it into Container C.}\\
       \texttt{Remove} & \textit{Remove the car from Box 1.} & \textit{Take the furby out of Container A.}\\
       \texttt{Put} & \textit{Put the car into Box 1.} & \textit{Place the furby inside Container A.}\\
       \bottomrule
   \end{tabular}
   }
   \caption{Different phrasings of the state-changing operations under the AltForms evaluation setup.}
   \label{tab:altforms-comparison}
\end{table*}

\paragraph{Demonstration/Test Mismatch for Form-meaning Disentanglement (AltForms)} As discussed in Sections~\ref{subsec:desiderata}--\ref{subsubsec:dataset}, we additionally experimented with a setup where the demonstration and test examples were mismatched in the form of the descriptions of the states and operations. Under this setup, the models were evaluated with set of object names and phrasings of the state and operation descriptions that were different from the demonstration examples (see Table~\ref{tab:altforms-comparison}). Except for the determiner \textit{the} and the preposition \textit{into}, the two sets share no words (although subwords may be shared depending on tokenization).

\paragraph{Scenarios with Greater Context Dependence (MoveContents, AmbiRef)} As also discussed in Sections~\ref{subsec:desiderata}--\ref{subsubsec:dataset}, we experimented with two additional scenarios with greater context dependence. The first scenario (MoveContents) introduces an additional operation \textit{Move contents of Box N to Box M} that moves all objects in Box N to Box M, that does not provide an explicit enumeration of objects being moved. This requires the model to rely on the preceding description to identify the set of objects being moved, further removing room for heuristics that allow the prediction of the final state without composing the initial description and the operations in temporal order. The second scenario (AmbiRef) adds adjectival modification to object names (e.g., \textit{the big brain} and \textit{the small brain}), where the adjective can be omitted in some of the operations, depending on the state of the box being described. Specifically, the modifier is dropped if there is only one object of a specific type in a box (e.g., \textit{Move the brain from Box 1 to Box 2} if there is only one brain in Box 1). When predicting the contents of a box, the model is instructed to output the object type with the correct adjective to fully disambiguate the referring expression for the object. This again requires composition of the initial description and the operations in temporal order to correctly interpret the otherwise ambiguous object mentions. (See Appendix~\ref{app:example} for example scenarios and operations.)

\begin{figure}[t]
    \centering
    \includegraphics[trim={0 0.55cm 0 0.15cm},clip,width=\columnwidth]{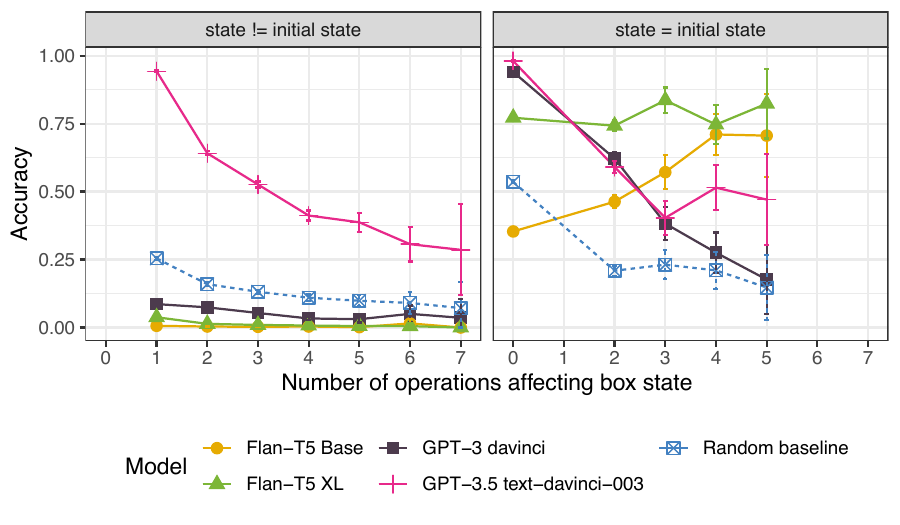}
    \caption{Accuracy on state prediction after $n$ operations that affect a specific box. Left: predictions for boxes whose content differs from the initial state, Right: predictions for boxes whose content is the same as in the initial state. Error bars show 95\% CIs. 
    }
    \label{fig:in-context-results}
\end{figure}

\subsection{Evaluation}
We estimated the entity tracking capacity of the models by computing the accuracy of predicting the contents of each box after each operation. Given that we rely on arbitrary-length cloze completion to predict the contents, we had to score unconstrained generations. While we only considered instances as correct where the output mentions all objects (and no additional objects) in a given box, our evaluation allowed for minor deviations from the exact form of the response. Namely, we allowed the objects to appear in any order, the object names may be separated by commas or \textit{and}, and both complete noun phrases with a determiner (e.g., \textit{the furby}) and bare nouns (\textit{e.g., furby}) are considered correct.

The task becomes intrinsically harder as more operations are applied to a specific box, since the initial state description needs to be combined sequentially with all subsequent operations. Further, similarly to our observation in Section~\ref{sec:motivation}, every operation changes the state of at most two boxes. This implies that the number of datapoints corresponding to fewer operations is much greater than the number of datapoints for more operations. For these reasons, we report accuracy as a function of the number of operations affecting a particular box rather than reporting aggregate accuracy, and show all results with 95\% confidence intervals.

\subsection{Results}
\label{subsec:results}

Figure~\ref{fig:in-context-results} shows the prediction accuracy for different number of operations that affected a box (e.g., 3 indicates that three operations changed the content of the box after the initial state). The left panel shows the instances where the probed state differed from the initial state; the right panel shows the instances where the probed state was the same as the initial state. As the left panel shows, only GPT-3.5 \texttt{text-davinci-003} consistently outperformed the (strong) random baseline. While, not surprisingly, the accuracy of this model also decreases as the number of operations increases, it still correctly predicted all contents of a box after 7 operations in more than 25\% of the cases. The Flan-T5 models, on the other hand, seemed to ignore the operations and primarily predicted the initial state description, as indicated by the consistently high accuracy when the final state matches the initial state (right panel), as well as the consistently low accuracy when the final state deviates from the initial state (left panel). GPT-3 \texttt{davinci} also primarily repeated the initial state, but as indicated by the steep decrease in the right panel, it was distracted by intervening operations even when repeating the initial state. 

\begin{figure}[h]
    \centering
    \includegraphics[trim={0 0.55cm 0 0.2cm},clip,width=\columnwidth]{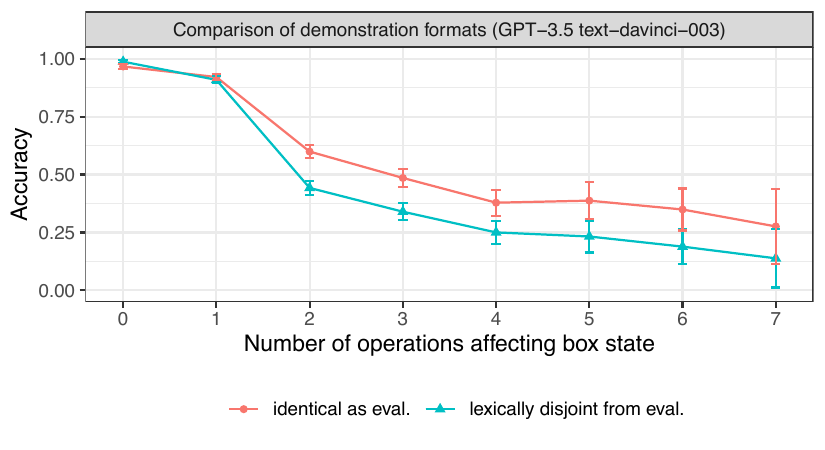}
    \caption{Entity tracking accuracy of \texttt{text-davinci-003} with low lexical overlap between demonstration and test examples (AltForms).}
    \label{fig:gpt3-base-altforms-results}
\end{figure}

\paragraph{Form-meaning Disentanglement} We additionally evaluated \texttt{text-davinci-003}, the only model that exhibited a non-trivial entity tracking capacity in the first set of results, under the AltForms setup as described in Section~\ref{subsec:prompting} where the demonstration examples have low lexical overlap with the test examples. Figure~\ref{fig:gpt3-base-altforms-results} shows the prediction accuracy of \texttt{text-davinci-003} on a representative subsample\footnote{For each number of operations $n$ affecting a box, we sampled 100 states with at least one example with $n$ operations.} of our data. The blue line represents the performance when the descriptions in the demonstration and the test examples have low lexical overlap. As the comparison to the original results (red line) shows, the disjoint demonstrations did lead to a small drop in performance when there were more than two operations affecting a box. Nevertheless, \texttt{text-davinci-003} was able to predict the correct state of entities in many cases, further adding support for its non-trivial entity tracking capacity.

\paragraph{Greater Context Dependence} Finally, we evaluated \texttt{text-davinci-003} on two additional datasets described in Section~\ref{subsec:prompting}: the AmbiRef dataset where adjectival modification can be omitted depending on the current context,  and the MoveContents dataset where a new operation \textit{Move contents of Box N to M} that requires the contents of Box N to be contextually identified.
\begin{figure}
    \centering
    \includegraphics[trim={0 0.55cm 0 0.2cm},clip,width=\columnwidth]{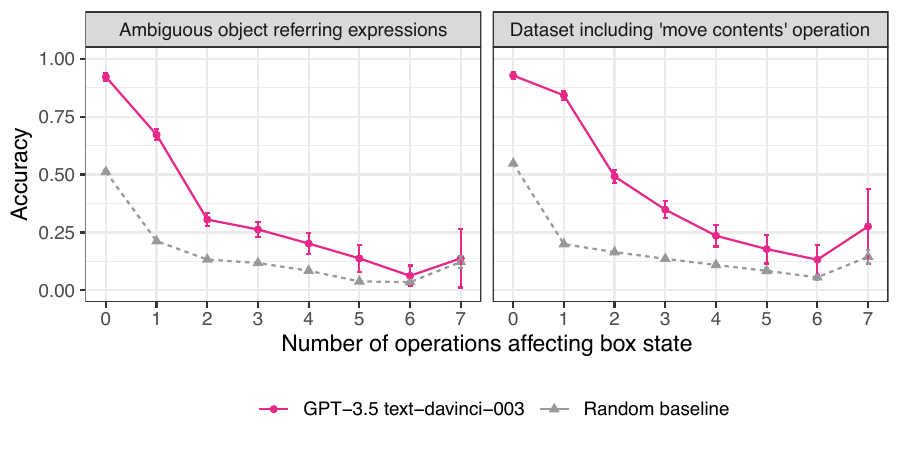}
    \caption{Entity tracking accuracy of \texttt{text-davinci-003} for the AmbiRef (left) and MoveContents (right) datasets. }
    \label{fig:gpt3-pragmatics-movecontents-results}
\end{figure}
Figure~\ref{fig:gpt3-pragmatics-movecontents-results} shows the performance of \texttt{text-davinci-003} on these two datasets compared to our random baseline. As these plots show, even in these more challenging cases, we observe non-trivial entity tracking abilities. However, as the number of operations grows, performance rapidly approaches the random baseline, suggesting that entity tracking becomes increasingly more brittle as more state changes need to be considered jointly.

\subsection{Discussion}
Our results show that among the models we evaluated, only GPT-3.5 \texttt{text-davinci-003} exhibit non-trivial entity tracking behavior. While its performance does decrease as the number of operations increases, the model still produced many accurate predictions even after six or seven sequences of operations. Furthermore, through the AltForms experiment with low lexical overlap between demonstration/test, we also ruled out the possibility that the demonstrations are teaching the model this task or that the model is primarily relying on superficial slot-filling heuristics. Therefore, we conclude that \texttt{text-davinci-003} does have some capacity to track discourse entities in linguistically expressed contexts. To a lesser extent, we also observed this capacity in the highly context-dependent AmbiRef and MoveContents experiments. This provides further evidence against superficial heuristics in the performance we observe; at the same time, this highlights scenarios in which even the most recent models exhibit difficulties.

On the other hand, entity tracking behavior did not surface in GPT-3 \texttt{davinci} (likely of similar size as GPT-3.5 \texttt{text-davinci-003}), a model pretrained primarily on text corpora on the next word prediction objective. This was also true for denoising models that have been finetuned on many tasks combined with instructions and demonstrations: the Flan-T5 models also showed near-zero accuracy on non-trivial examples.
\begin{figure}
    \centering
    \includegraphics[trim={0 0.55cm 0 0.15cm},clip,width=\columnwidth]{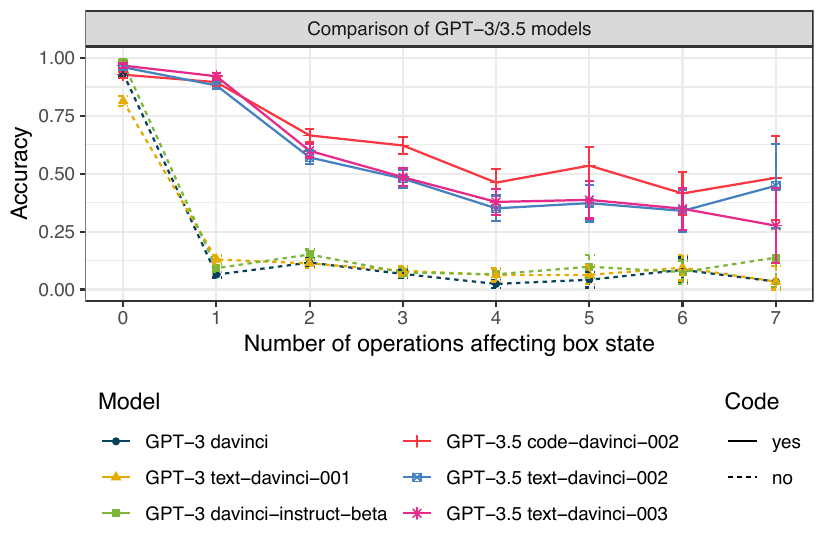}
    \caption{Accuracy on state prediction for different GPT-3 models. Solid lines denote models trained on code and text, and dotted lines denote models mainly trained on text.
    }
    \label{fig:gpt-comparisons}
\end{figure}

Our results overall show that there exists a language model that can perform entity tracking to some degree, but this capacity does not necessarily surface in all sufficiently large models trained on large corpora. Then, which factors are responsible for this difference? Given that \texttt{davinci} and \texttt{text-davinci-003} differ along at least two dimensions (\texttt{text-davinci-003} is based on a model that was trained on code and it was trained with additional human feedback \citep{ouyang2022training}; see Table~\ref{tab:in-context-models}), our initial results do not shed light on what exactly contributes to this difference. We therefore conducted a follow-up experiment where we compared a range of GPT-3 and GPT-3.5 models to identify factors that contribute to the stark difference between \texttt{davinci} and \texttt{text-davinci-003}.\footnote{To limit inference costs, we used the same subsample of data as in the AltForms experiment.} 

\begin{figure*}
    \centering
    \includegraphics[trim={0 0.45cm 0 0.15cm},clip,width=\textwidth]{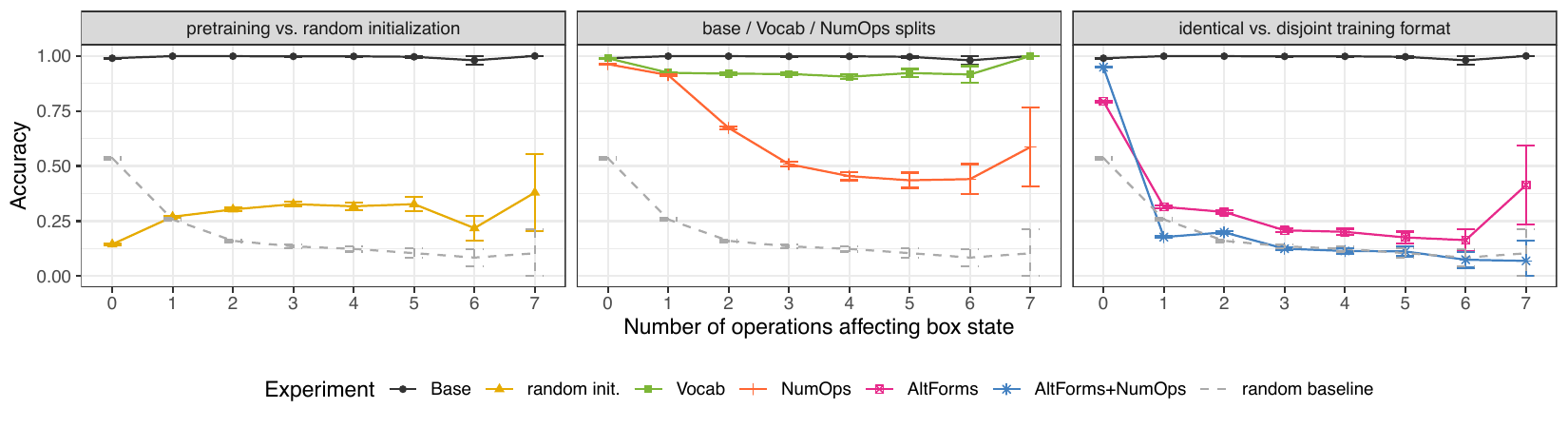}
    \caption{
    Results for finetuned T5 models.}
    \label{fig:base-random-nofinetune-results}
\end{figure*}

\paragraph{Training on Code Encourages Entity Tracking Behavior} As Table~\ref{tab:in-context-models} shows, two key dimensions of variation across models are additional training on human feedback and pretraining on code. If additional training on human feedback imbues language models with the ability to track entities, all models except for GPT-3 \texttt{davinci} and GPT-3.5 \texttt{code-davinci-002} should be able to track entities. If, on the other hand, pretraining on code leads to better entity tracking, we expect all GPT-3.5 models to outperform GPT-3 on our task. As Figure~\ref{fig:gpt-comparisons} shows, GPT-3.5 models that have been trained on code systematically outperformed GPT-3 models, including \texttt{code-davinci-002} that was not trained on human feedback. This suggests that a substantial representation of code in the pretraining data is beneficial for a language model's entity tracking capacity to surface. 

A further question that our results so far do not answer is to what extent model size matters and whether models at the scale of Flan-T5 can also exhibit non-trivial entity tracking behavior. Since there exist no smaller models that have been trained with the same objective and training data as the GPT-3.5 models, we explore this question through finetuning experiments with T5.

\section{Experiment 2: Finetuning}

We investigated whether smaller models at the scale of T5 can \textit{learn} to track entity states through a series of experiments where we provide supervised training to the models.

\subsection{Train/test splits}
\label{subsec:splits}

As discussed in Section~\ref{subsec:desiderata}, one challenge of evaluating entity tracking abilities is distinguishing this capacity from simple heuristics such as templatic slot-filling.
We therefore designed various types of training/evaluation mismatches that block several possible shortcuts as described below.

\vspace{0.15em}\noindent \textbf{Base Split}
Here, we used the same format for training and evaluation examples. All initial states differed across training and evaluation to block simple slot-filling heuristics, as discussed in Section~\ref{subsubsec:dataset}.

\vspace{0.15em}\noindent \textbf{NumOps Split}
The NumOps split restricts the maximum number of operations within a single example in the training set to 2, but includes up to 12 operations in the evaluation set. This split was intended to test whether a finetuned model is able to generalize to longer sequences of operations than it has seen during finetuning.

\vspace{0.15em}\noindent \textbf{Vocab/AltForms Splits}
The vocab split tests whether objects that are not part of the set of objects used during training can also be adequately tracked. We compiled a list of comparatively infrequent object names (e.g., \textit{pomelo, furby, Flav-R-Straw}; not in BNC) and sampled the training and test sets using two completely disjoint sets of object names. The training set used the infrequent object list and the test set used the original object list. The AltForm split follows the design described in Section~\ref{subsec:prompting}. These splits aim to tease apart whether the model learns to associate specific words/phrases with the operations or whether finetuning leads to more generalizable entity tracking behavior. 

\vspace{0.15em}\noindent \textbf{AmbiRef/MoveContents Splits}
The AmbiRef and MoveContents splits follow the design described in Section~\ref{subsec:prompting}. These splits aim to test whether the model can learn to interpret operations that are underspecified without considering the current state of the affected boxes. For these splits, the training and test examples share the same format.

\subsection{Models}
We evaluated T5-base, the best-performing model in \citet{li-etal-2021-implicit}, by finetuning it on each of the datasets described above.
As an additional baseline, we compared against T5 with randomly initialized parameters trained directly on our datasets. 

\subsection{Results and Discussion}

\noindent \paragraph{Pretrained T5 can Learn to Perform Entity Tracking} As shown in Figure~\ref{fig:base-random-nofinetune-results} (left), finetuning T5 leads to near-perfect accuracy on the base split. This suggests that the model is capable of learning this task. Training a randomly initialized T5 did not yield the same result: the accuracy of a model trained from random weights is considerably lower, due to the model almost exclusively predicting that a box is empty. These two results suggest that pretraining is crucial for the model to be able to learn this task.
Furthermore, the model's entity tracking capacity is robust to novel object names at test time, with only minor degradation on accuracy (Figure~\ref{fig:base-random-nofinetune-results}, middle). Training only on operation sequences with a maximum length of 2 (NumOps split) leads to a larger degradation in performance for longer operation sequences, but even for longer operation sequences, the model is able to infer the correct final state in more than 45\% of the cases. Finally, the model performance does degrade substantially when the training examples have low lexical overlap with test examples (Figure~\ref{fig:base-random-nofinetune-results}, right). Nevertheless, model performance remains above the random baseline when the model is trained on up to 12 operations (pink line). If we trained only on up to two operations (blue line), however, the performance degradation was compounded and performance no longer exceeded the random baseline. These results suggest that finetuning on an entity tracking task does lead to entity tracking abilities that generalize to many challenging scenarios. At the same time, however, performance rapidly degrades as the finetuning and evaluation splits become increasingly dissimilar, and it remains an open question to what extent finetuning on a limited domain such as the boxes environment transfers to more general entity tracking abilities in more naturalistic discourse.
\begin{figure}
    \centering
    \includegraphics[trim={0 0.55cm 0 0.2cm},clip,width=\columnwidth]{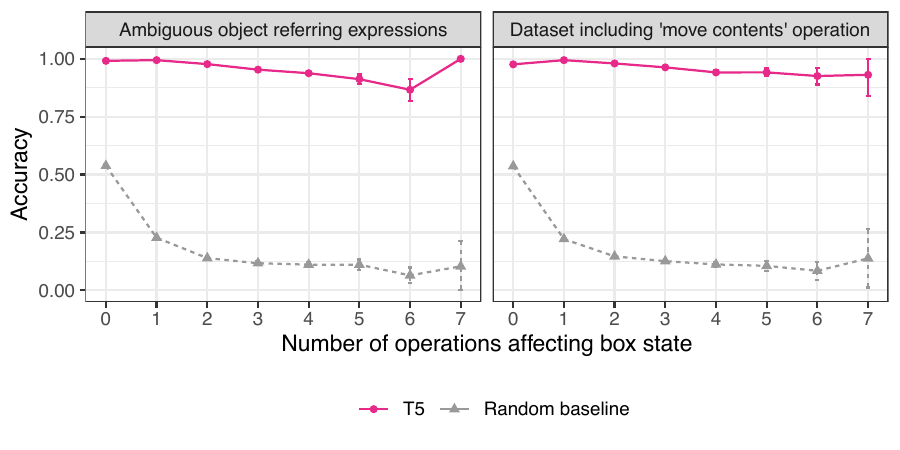}
    \caption{Accuracy of finetuned T5 for AmbiRef (left) and MoveContents (right) datasets.}
    \label{fig:t5-pragmatics-movecontents-results}
\end{figure}

\paragraph{Interpreting Context-Dependent Operations Remains Challenging} If we include operations that either contain ambiguous referring expressions (Figure~\ref{fig:t5-pragmatics-movecontents-results}, left) or a \textit{Move contents} operation (Figure~\ref{fig:t5-pragmatics-movecontents-results}, right) in the finetuning and evaluation data, we see a slight degradation in performance and the model no longer achieves near-perfect accuracy, despite training and evaluating on examples of the same format. In both cases, the model almost exclusively made mistakes with examples that require the interpretation of context-dependent operations. This suggests that the context-dependent operations do compound the difficulty of entity tracking even when the models receive direct training signal.

\section{Conclusion}

We set out to investigate whether pretrained LLMs exhibit entity tracking behavior. We developed a task that allowed us to evaluate whether LLMs can predict the state of an entity based on an initial state description and operations that act upon it. In the first set of experiments, we found that GPT-3 \texttt{davinci}, a vanilla pretrained language model, and Flan-T5, an instruction-finetuned language model, fail at this task and simply repeat the initial state description. The GPT-3.5 models (the core difference with the aforementioned models being code in pretraining corpora), on the other hand, exhibited non-trivial entity tracking behavior. Even after many operations affecting an entity state, they consistently performed above a strong random baseline. In the second set of experiments, we showed that this behavior can also to a large extent be learned by smaller models such as T5. When we finetune the model on this task, it is also able to perform entity tracking on examples that differ along several dimensions from the training data. 
Taken together, our results provide evidence that (a) vanilla language models that have mainly been trained on text data do not exhibit entity tracking abilities out of the box, (b) pretraining on code and text data considerably improves this ability,\footnote{This is also compatible with the observation by \citet{sap-etal-2022-neural} that GPT-3.5 performs considerably better than GPT-3 in answering factual questions about object location changes. Further, see \newcite{madaan-etal-2022-language} for other tasks for which pretraining on code seems to be beneficial.} and (c) finetuning on this task can make this behavior also surface in smaller models that have primarily been trained on text data, although it remains an open question how general this ability is in smaller finetuned models.

What are reasons behind the efficacy of training both on text and code? For producing correct code, tracking the states of variables is important. Therefore, to speculate, this kind of pretraining data may provide a stronger signal for the model to track entities compared to pure text data. It could also be that, as previously speculated \cite[][i.a.]{potts2020possible,merrill-etal-2021-provable}, pretraining on code provides additional grounding. 

The present results also highlight the importance of the composition of the pretraining data. Our results showed that models that are trained on the language modeling objective on large corpora show dramatically different behavior on entity tracking depending on whether the training data includes substantial amount of code or not.
In this light, future work could investigate the effect of pretraining on code more widely, including its effect on the model representations and to what extent it facilitates the emergence of world models in LMs.

Finally, we also laid out several principles that should be followed when evaluating state tracking abilities. Apart from these specific principles, we make a more general point that in assessing abilities related to meaning, one needs to consider potential strategies that the model could use to solve the task and make sure that the test examples do not mimic the distributional patterns of the training or finetuning data. 
Only then can we properly assess meaning-related capacities of LMs.

\section*{Limitations}

One limitation of this work is that we are only considering behavioral data which makes it difficult to establish a fully causal link between entity tracking capacities and high performance on our task. Entity tracking is a high-level linguistic behavior and many other capacities are necessary for achieving high accuracy on our task. Therefore, we cannot rule out that differences in some other capacity, such as interpreting sentences compositionally (see \citealt{bogin-2022-tweet} and \citealt{bogin-etal-2022-unobserved} for evidence that GPT-3 and GPT-3.5 models differ in their compositional generalization behavior), are the main driver for the differences in behavior we see across models. 

A possible criticism of our setup is that it requires short-term memory capacities that exceed the memory capacities of most, if not all, humans. That is, if we presented humans with the same input as the model, we would not expect them to be able to keep track of the contents of all 7 boxes due to memory limitations. Therefore we are potentially expecting models to do super-human entity tracking, a setup that has been criticized for model evaluations of other linguistic abilities \cite{lampinen2022can}. We nevertheless believe that our task is justified given the architecture of the evaluated models. Transformer-based models can look back to any token in the entire input sequence within their context window, so a proper comparison between humans and models would be to present humans with the full description in written form and let them re-read the description after being prompted to state the contents of a box. While we did not formally evaluate whether humans have this ability on a larger population, we personally did not have any trouble tracking the contents of boxes when we had access to the written description.

Relatedly, we designed our task such that the entire description fits within the context window of pretrained language models. However, as we mentioned in the introduction, entity tracking is an important ability for understanding long contexts and given the limited context window, our results do not apply to texts whose length exceeds a model's context window, and likely different model architectures will be necessary to perform proper entity tracking for longer texts.

Further, while we found that GPT-3.5 models as well as finetuned T5 models can track entities in our task with higher accuracy than a strong random baseline, our results also indicate that this behavior is not very stable once several operations act on an entity. Our results should therefore not be taken as justification for using these models for critical applications where high accuracy is needed.

Lastly, we only evaluated English models in this work. Given that we showed that even without high  lexical overlap between the training and evaluation examples, models can keep track of entities to some extent, it seems likely that our results also apply to other languages. However, whether this actually the case remains an open question.

\section*{Acknowledgements}
We thank Jacob Andreas, Ellie Pavlick, Allyson Ettinger, Tal Linzen, Will Merrill, and the members of the NYU Computation and Psycholinguistics lab for discussions, and Belinda Li for sharing model outputs and details about their data preparation procedures and experiments. We thank Cookie for contributing to the authorship decision and for emotional support. This research was conducted in part through the NYU IT High Performance Computing resources, services, and staff expertise, and it
 was supported by the NSF under Grant \#2030859 to the
Computing Research Association for the CIFellows Project and the European Research Council (ERC) under the European Union's Horizon 2020 Research and Innovation Program (Grant Agreement \#948878). Any opinions, findings, and conclusions or recommendations expressed in this material are those of the authors and do not necessarily reflect
the views of the National Science Foundation nor the Computing Research Association.

\bibliography{anthology,custom}
\bibliographystyle{acl_natbib}

\newpage
\appendix

\section{Additional Analyses of Results from \citet{li-etal-2021-implicit}}
\label{app:implicit-meaning}

As mentioned in Footnote~\ref{footnote:li}, \newcite{li-etal-2021-implicit} conducted two more experiments that prima facie provided additional evidence for implicit meaning representations and state tracking abilities in language models. However, the data and setup of these two experiments also likely overestimates models' abilities.

In the second set of probing classifier experiments, \newcite{li-etal-2021-implicit} used data generated using the TextWorld engine \cite{cote2019textworld}. In this setup, there is a textual description of several entities in a text-based game (e.g., \textit{a wooden door}) as well as actions that a player took (e.g., \textit{opening the wooden door}). Their probing classifier takes the representations of either one entity and a property of that entity (e.g., that \textit{the wooden door is \textbf{closed}}) or the representations of two entities and a relation between them (e.g., that the \textit{king-sized bed \textbf{is in} the bedroom}) and from these representations, the classifier has to predict whether a given proposition is true or false considering the initial description and the series of actions that a player took. Only propositions that involve entities that have been mentioned are probed. The issue with this setup is that there are many propositions that are always true in both the training and evaluation splits (e.g., in all the game simulations, the chest drawer is in the bedroom, so the probing classifier should always return true for this input independent of the previous context). Furthermore, even for the entity-property propositions and entity-relation-entity propositions which are not always true in the training and evaluation data, the data is very biased and a baseline that predicts the most common answer in the training data without taking the initial descriptions or the user actions into account (a violation of Desideratum 3, see Section~\ref{subsec:desiderata}), already achieves an accuracy of 88.5\%, a number that puts the reported probing classifier accuracy of 96.9\% into a bit more context.

Further, \newcite{li-etal-2021-implicit}, also presented an experiment where they manipulated specific entity representations of a synthetic version of the Alchemy dataset. In this experiment, they first encoded an initial description $D$ and an operation $O$ which affected a beaker $X$ using a language model that had been finetuned to predict the next operation, resulting in representation $R_1$. Then, they encoded the same initial representation $D$ and an operation of the form \textit{Drain $n$ from beaker $Y$}, where beaker $Y$ was always different from beaker $X$ and $n$ was the amount of liquid that was in beaker $Y$ according to the initial description. This resulted in representation $R_2$. Then, they extracted the representation of the initial description of beaker $Y$ from representation $R_2$ and replaced the representation of beaker $Y$ in $R_1$ with the corresponding representation in $R_2$ to obtain $R_{mixed}$. They then used $R_{mixed}$ as an input to T5 and showed that the predicted next operation based on $R_{mixed}$ was considerably more often compatible with both operations (the operation encoded in $R_1$ and the operation encoded in $R_2$) compared to predicting the next operation from $R_1$ or $R_2$, which they took as evidence that the token representations of the initial description encoded the entities state after performing the operation. The issue with this experiment is that the operation from which $R_2$ was computed was always of the form \textit{Drain $n$ from $Y$th beaker}, so the final state of beaker $Y$ was always empty. Therefore, this experiment primarily shows that the token \textit{drain} affected the representation of the initial state (and subsequently the prediction of the next operation) but not more generally, that the actual state of the manipulated beaker $Y$ is fully encoded in its initial state description. To answer that question, one would have to repeat this experiment with more complex operations that do not give away the final state (a violation of Desideratum 2, see Section~\ref{subsec:desiderata}).

In summary, the additional experiments in \newcite{li-etal-2021-implicit} also violate some of the desiderata we laid out in Section~\ref{subsec:desiderata}, and therefore it is difficult to draw conclusions about state tracking abilities from these experiments.

\section{Example Input-Output Pairs}
\label{app:example}

\ref{ex:in-out} shows an example input-output pair from our base dataset (NumOps on Box 6 = 2).

\ex. \label{ex:in-out}
    \a. \textsc{Input:} \textit{Box 0 contains the painting, Box 1 contains the bell, Box 2 contains the guitar, Box 3 contains the egg and the mirror and the sheet, Box 4 contains the chemical, Box 5 contains the disk and the wire, Box 6 contains the glass and the knife. Move the glass from Box 6 to Box 4. Put the gift into Box 5. Move the guitar from Box 2 to Box 6. Put the milk into Box 4. Remove the mirror and the sheet from Box 3. Box 6 \_\_}
    \b. \textsc{Output:} \textit{contains the guitar and the knife.}

\noindent \ref{ex:in-out-pragmatics} shows an example input-output pair from the dataset with operations that contain referring expressions that are ambiguous without considering the initial state and the previous operations (NumOps on Box 6 = 2). Note that in order to correctly interpret \textit{Move the guitar from Box 2 to Box 6}, the model has to consider the information that the blue guitar (as opposed to the red guitar) was in Box 2 prior to the operation. Hence, this operation is ambiguous out of context.

\ex. \label{ex:in-out-pragmatics}
    \a. \textsc{Input:} \textit{Box 0 contains the yellow book and the green flower and the red guitar, Box 1 contains the small bomb and the small book and the blue bone, Box 2 contains the blue guitar, Box 3 contains the blue bell, Box 4 contains the green paper and the yellow note and the yellow television, Box 5 contains the yellow bell, Box 6 is empty. Move the guitar from Box 2 to Box 6. Put the blue wire and the big television into Box 5. Move the flower from Box 0 to Box 6. Box 6 \_\_}
    \b. \textsc{Output:} \textit{contains the blue guitar and the green flower.}

\ref{ex:in-out-movecontents} shows an example input-output pair from the dataset with the \textit{Move contents of} operation (NumOps on Box 6 = 2). Note that in order to correctly interpret \textit{Move the contents of Box 2 to Box 6}, the model has to consider the information that the tea (as opposed to the red guitar) was in Box 2 prior to the operation.

\ex. \label{ex:in-out-movecontents}
    \a. \textsc{Input:} \textit{Box 0 contains the fan and the gift and the letter, Box 1 contains the beer and the mirror and the tie, Box 2 contains the tea, Box 3 contains the boot, Box 4 contains the coat and the plate and the shirt, Box 5 contains the bottle, Box 6 is empty. Move the contents of Box 2 to Box 6. Put the dress and the painting into Box 5. Move the letter from Box 0 to Box 6. Box 6  \_\_}
    \b. \textsc{Output:} \textit{contains the letter and the tea.}

\begin{figure*}[t]
    \centering
    \includegraphics[trim={0 0.55cm 0 0.2cm},clip,width=\textwidth]{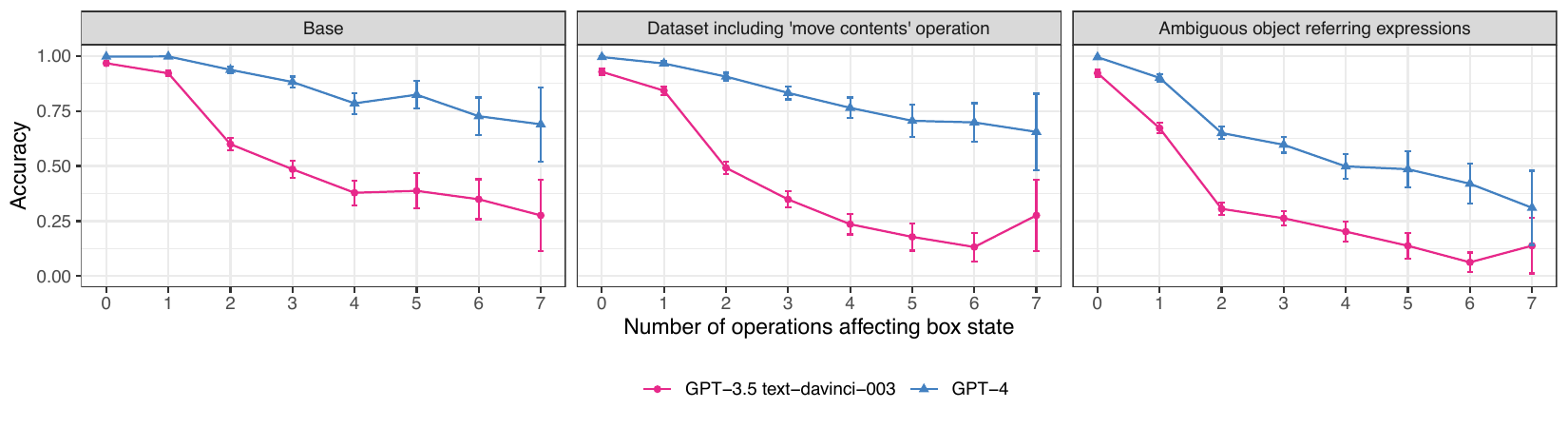}
    \caption{Entity tracking accuracy of \texttt{text-davinci-003} and GPT-4 in 2-shot in-context learning experiments.}
    \label{fig:gpt4-results}
\end{figure*}

\section{Implementation Details}
\label{app:model-details}

\vspace{0.15em}\noindent \textbf{In-context} For GPT-3, we used the OpenAI API. We used greedy decoding with the temperature parameter set to 0, used a maximum target generation length of 150, and used the newline character as an additional stop token. Inference time and model size is not available for GPT-3, but we generated about 16 million tokens in total. All GPT-3 experiments that we report here in the paper were conducted in January 2023.

For Flan-T5, we used a beam size of 3 and a maximum target generation length of 256. Other hyperparameters were kept as the default values of \texttt{T5ConditionalGeneration} in the HuggingFace library. Inference for T5-XL took about 5 hours on a single A100 GPU, and for T5 base, about 2 hours.

\vspace{0.15em}\noindent \textbf{Finetuning} We finetuned T5 for a single epoch using a batch size of 8 and a learning rate of $1 \times 10^{-4}$. In initial explorations, increasing the number of finetuning epochs did not yield substantial gains on development set performance, and was sometimes even harmful. Training and inference took around 3 hours on a single RTX8000 GPU.

\section{Additional Results}
\label{app:additional-results}
\subsection{GPT-3.5 Zero-shot Results}
As mentioned in the main text, GPT-3.5 was able to to output the contents of boxes in the correct format without any in-context demonstrations (see Table~\ref{tab:zero-shot-prompt} for the prompt template).

\begin{figure}[h]
    \centering
    \includegraphics[trim={0 0.55cm 0 0.2cm},clip,width=\columnwidth]{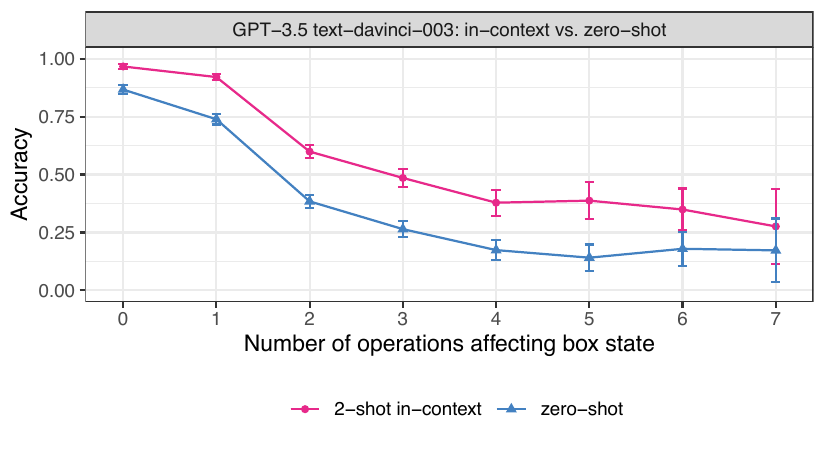}
    \caption{Entity tracking accuracy of GPT 3.5 under in-context (2-shot) and zero-shot settings.}
    \label{fig:gpt3-zero-shot}
\end{figure}

Figure~\ref{fig:gpt3-zero-shot} compares the performance of \texttt{text-davinci-003} in the 2-shot setting to the zero-shot setting. Model performance degraded slightly without demonstration examples, which suggests that the examples are indeed helpful in guiding the model to correctly perform the task. Nevertheless, we observed non-negligible performance even in the zero-shot setting, which corroborates the conclusion that GPT 3.5 exhibits non-trivial entity tracking capacities.

\subsection{GPT-4 Results}
We also probed the more recently released GPT-4 model \cite{gpt4} through the OpenAI API. Figure~\ref{fig:gpt4-results} compares the performance of GPT-3.5 \texttt{text-davinci-003} and GPT-4 on the base dataset (left) as well as the more challenging MoveContents (middle) and AmbiRef (right) datasets discussed in Section~\ref{subsec:prompting}. Across all datasets, GPT-4 performed considerably better than GPT-3.5. But similarly to GPT-3.5, performance degraded as more operations affected a box. Considering that no public information about the training data, training procedure or architectural details about GPT-4 exists, we cannot draw any conclusions about the reason behind the improved performance of GPT-4 compared to previous models.

\section{Prompts}
\label{app:prompts}

Table~\ref{tab:in-context-prompts} shows the 2-shot prompts used for in-context experiments. Table~\ref{tab:in-context-prompts-disjoint} shows the 2-shot prompts used for the AltForms experiments, where the demonstrations contain descriptions that have lower lexical overlap with the test examples. Table~\ref{tab:zero-shot-prompt} shows the zero-shot prompts discussed in Appendix~\ref{app:additional-results}. All prompts are available at \url{https://github.com/sebschu/entity-tracking-lms}.

\begin{table*}
\begin{tabular}{p{0.9\linewidth}}
    \toprule
    \textbf{2-shot prompt with all boxes queried at once (GPT-3 experiments)} \\
    \midrule
    Given the description after "Description:", write a true statement about all boxes and their contents to the description after "Statement:".\\\\
    
    Description: Box 0 contains the car, Box 1 contains the cross, Box 2 contains the bag and the machine, Box 3 contains the paper and the string, Box 4 contains the bill, Box 5 contains the apple and the cash and the glass, Box 6 contains the bottle and the map.\\
    Statement: Box 0 contains the car, Box 1 contains the cross, Box 2 contains the bag and the machine, Box 3 contains the paper and the string, Box 4 contains the bill, Box 5 contains the apple and the cash and the glass, Box 6 contains the bottle and the map.\\\\
    
    Description: Box 0 contains the car, Box 1 contains the cross, Box 2 contains the bag and the machine, Box 3 contains the paper and the string, Box 4 contains the bill, Box 5 contains the apple and the cash and the glass, Box 6 contains the bottle and the map. Remove the car from Box 0. Remove the paper and the string from Box 3. Put the plane into Box 0. Move the map from Box 6 to Box 2. Remove the bill from Box 4. Put the coat into Box 3.\\
    Statement: Box 0 contains the plane, Box 1 contains the cross, Box 2 contains the bag and the machine and the map, Box 3 contains the coat, Box 4 contains nothing, Box 5 contains the apple and the cash and the glass, Box 6 contains the bottle.\\\\
    
    Description: \{\texttt{description}\}\\
    Statement: Box 0 contains\\   
    \midrule
    \textbf{2-shot prompt with boxes queried individually (T5 experiments)} \\
    \midrule
    Given the description after "Description:", write a true statement about a box and the contents of this box according to the description after "Statement:".\\\\
    
    Description: Box 0 contains the car, Box 1 contains the cross, Box 2 contains the bag and the machine, Box 3 contains the paper and the string, Box 4 contains the bill, Box 5 contains the apple and the cash and the glass, Box 6 contains the bottle and the map.\\
    Statement: Box 1 contains the cross.\\\\
    
    Description: Box 0 contains the car, Box 1 contains the cross, Box 2 contains the bag and the machine, Box 3 contains the paper and the string, Box 4 contains the bill, Box 5 contains the apple and the cash and the glass, Box 6 contains the bottle and the map. Remove the car from Box 0. Remove the paper and the string from Box 3. Put the plane into Box 0. Move the map from Box 6 to Box 2. Remove the bill from Box 4. Put the coat into Box 3.\\
    Statement: Box 2 contains the bag and the machine and the map.\\\\
    
    Description: \{\texttt{description}\}\\
    Statement: Box \{\texttt{boxnum}\} contains\\
    \bottomrule
\end{tabular}
\caption{Prompts with 2-shot in-context demonstrations.}
\label{tab:in-context-prompts}
\end{table*}

\begin{table*}
\begin{tabular}{p{0.9\linewidth}}
    \toprule
    \textbf{2-shot prompt with disjoint surface forms from test examples} \\
    \midrule
    Given the description after "Description:", write a true statement about all containers or boxes and their contents to the description after "Statement:".\\\\
    
    Description: The biscotti is in Container A, the icicle is in Container B, the granite and the machine are in Container C, the folio and the encyclopedia are in Container D, the bill is in Container E, the spork and the jackknife and the frappuccino are in Container F, the clipper and the ladybug are in Container G.\\
    Statement: Container A contains the biscotti, Container B contains the icicle, Container C contains the granite and the machine, Container D contains the folio and the encyclopedia, Container E contains the bill, Container F contains the spork and the jackknife and the frappuccino, Container G contains the clipper and the ladybug.\\\\
    
    Description: The biscotti is in Container A, the icicle is in Container B, the granite and the machine are in Container C, the folio and the encyclopedia are in Container D, the bill is in Container E, the spork and the jackknife and the frappuccino are in Container F, the clipper and the ladybug are in Container G. Take the biscotti out of Container A. Take the folio and the encyclopedia out of container D. Place the tetrapod inside Container A. Pick up the ladybug in Container G and place it into Container C. Take the bill out of Container E. Place the gumball inside Container D.\\
    Statement: Container A contains the tetrapod, Container B contains the icicle, Container C contains the granite and the machine and the ladybug, Container D contains the gumball, Container E contains nothing, Container F contains the spork and the jackknife and the frappuccino, Container G contains the clipper.\\\\

    Description: \{\texttt{description}\}\\
    Statement: Box 0 contains\\
    \bottomrule
\end{tabular}
\caption{Prompt with 2-shot in-context demonstrations that have disjoint surface forms.}
\label{tab:in-context-prompts-disjoint}
\end{table*}

\begin{table*}
\begin{tabular}{p{0.9\linewidth}}
    \toprule
    \textbf{Zero-shot prompt} \\
    \midrule
    Given the description after "Description:", write a true statement about all boxes and their contents according to the description after "Statement:". Format the statement as Box 0 contains the A, Box 1 contains the B, Box 2 contains the C, Box 3 contains the D, Box 4 contains the E, Box 5 contains the F, Box 6 contains the G. A, B, C, D, E, F, G are placeholders for the contents of each box.\\\\
    
    Description: \{\texttt{description}\}\\
    Statement: Box 0 contains\\
    \bottomrule
\end{tabular}
\caption{Prompt template for zero-shot experiments.}
\label{tab:zero-shot-prompt}
\end{table*}

\section{Dataset Statistics and License Information}

See Tables~\ref{tab:dataset-statistics-exp1} and \ref{tab:dataset-statistics-exp2} for descriptive statistics of our datasets. All datasets are released under the GNU General Public License v3.0.

\begin{table*}[t]
\centering

    \begin{tabular}{l | cc | cc}
        &  \multicolumn{2}{ c | }{\textbf{Scenarios}}  &  \multicolumn{2}{ c  }{\textbf{Examples}}  \\
        \textbf{Dataset} & \textbf{Demonstration}  & 
        \textbf{Test}  & 
        \textbf{Demonstration}  & 
        \textbf{Test}   \\\midrule
        Complete & 1 & 990  & 14 &  90,090      \\
        Subsample & 1 & 491 & 14 & 5,012\\
        AmbiRef. & 1 & 474 & 14 & 4,991 \\
        MoveContents & 1 & 451 & 14 & 4,956\\
        
        \bottomrule
    \end{tabular}
    
    \caption{Dataset statistics for Experiment 1.}
    \label{tab:dataset-statistics-exp1}
\end{table*}

\begin{table*}[t]
\centering

    \begin{tabular}{l | ccc | ccc}
        & & \textbf{Scenarios} & & &  \textbf{Examples} &  \\
        \textbf{Dataset} & \textbf{Train}  & 
        \textbf{Dev}  & 
        \textbf{Test}  & 
        \textbf{Train}  & 
        \textbf{Dev} & 
        \textbf{Test}   \\\midrule
        Base & 990 & 220 & 990 & 90,090 & 20,020 & 90,090      \\
        NumOps & 990 & 220 & 990 & 20,790 & 20,020 & 90,090\\
        Vocab & 990 & 220 & 990 & 90,090 & 20,020 & 90,090  \\
        AltForms &  990 & 220 & 990 & 90,090 & 20,020 & 90,090  \\
        AltForms+NumOps &  990 & 220 & 990 & 20,790 & 20,020 & 90,090  \\
        AmbiRef &  990 & 220 & 990 & 90,090 & 20,020 & 90,090  \\
        MoveContents &  990 & 220 & 990 & 90,090 & 20,020 & 90,090  \\

        \bottomrule
    \end{tabular}
    
    \caption{Dataset statistics for Experiment 2.}
    \label{tab:dataset-statistics-exp2}
\end{table*}

\end{document}